\documentclass[11pt]{article}

\usepackage{acl}

\usepackage{times}
\usepackage{latexsym}
\usepackage[T1]{fontenc}
\usepackage[utf8]{inputenc}
\usepackage{microtype}
\usepackage[table]{xcolor}
\usepackage{graphicx}
\usepackage{booktabs}
\usepackage{amsmath}
\newcommand{\stage}[1]{%
    \raisebox{.12ex}{\textcircled{\raisebox{-.15ex}{\scriptsize #1}}}%
}
\title{
GroundSight at GroundLM 2026 Shared Tasks: GoldenViewVQA
}

\author{
  \textbf{Kun Wang\textsuperscript{1,3}},
  \textbf{Yupeng Hu\textsuperscript{1}},
  \textbf{Ruping Cao\textsuperscript{1}},
  \textbf{Hao Liu\textsuperscript{1}},
  \textbf{Zhiran Li\textsuperscript{1}} \\
  \textbf{Qianlong Xiang\textsuperscript{2}},
  \textbf{Harry Cheng\textsuperscript{3}} \\
  \textsuperscript{1}School of Software, Shandong University, Jinan, China \\
  \textsuperscript{2}School of Computer Science and Technology, \\
  Harbin Institute of Technology (Shenzhen), Shenzhen, China \\
  \textsuperscript{3}School of Computing, National University of Singapore, Singapore \\
  \texttt{khylon.kun.wang@gmail.com} \quad
  \texttt{huyupeng@sdu.edu.cn} \\
  \texttt{caoruping657@gmail.com} \quad
  \texttt{liuh90210@gmail.com} \\
  \texttt{zhiranli325@gmail.com} \quad
  \texttt{xiangqianlongcs@gmail.com} \\
  \texttt{xacheng1996@gmail.com}
}
\begin{document}

\maketitle

\begin{abstract}

GoldenViewVQA requires models to jointly answer driving-scene questions and identify the camera view containing the supporting visual evidence, making precise evidence localization as important as answer correctness. We present \textbf{CoVeR-VQA}, a training-free multi-stage verification and correction framework for grounded multi-view VQA. Starting from GPT-5.6 zero-shot predictions, CoVeR-VQA progressively applies view-specific verification with Gemini-3.6-Flash, prior-guided joint verification with Claude-Opus-5, and cross-split group-level verification that exploits semantically filtered question groups from shared multi-view scenes and validation-derived prior knowledge. On the official GoldenViewVQA test set, the four-stage CoVeR-VQA pipeline achieves 84.75\% Joint Accuracy, improving the GPT-5.6 zero-shot baseline by 13.56 percentage points, while reaching 94.92\% Answer Accuracy and 86.44\% View Accuracy. The final submitted run achieves 88.14\% Joint Accuracy after two additional evaluator-informed post-hoc corrections. Our analysis shows that supporting-view localization remains the primary source of residual errors, highlighting the importance of explicit evidence verification for reliable multi-view multimodal reasoning.

\end{abstract}

\section{Introduction}

Multimodal large language models (MLLMs) have demonstrated strong capabilities in visual representation and understanding, visual question answering, cross-modal retrieval and multimodal evaluation~\citep{bai2025qwen3vl,liu2025curmim,wang2025redundancy,li2026unim}. Visual question answering has also been increasingly explored in autonomous-driving scenarios~\citep{qian2024nuscenesqa}; however, producing a correct answer alone is insufficient for reliable grounded reasoning. A reliable system should also identify the visual evidence that supports its prediction. This requirement is particularly challenging in multi-view settings, where a scene is simultaneously observed from multiple cameras with different fields of view~\citep{yeh2026seeing}. Relevant evidence may appear in only one view or be ambiguous across several cameras, making precise evidence localization substantially harder than answer prediction itself~\citep{hu2021video,hu2021coarse,hu2023semantic,wang2024explicit,liu2025gaming,wang2026cross,hu2026visual}.

GroundLM 2026 comprises two shared tasks, GoldenViewVQA~\citep{wang2026doesanswercomefrom} and LitTraceQA~\citep{liu2026littraceqabenchmarkmultistagegrounding}; the overall task setup and findings are summarized in the shared-task findings paper~\citep{wang-etal-2026-groundlm}. In this work, we focus on GoldenViewVQA, which explicitly evaluates view-level visual evidence identification in multi-view autonomous-driving scenarios. Each instance contains six synchronized driving-scene camera views and a multiple-choice question, and systems are required to predict both the correct answer and its supporting camera view. The primary metric, Joint Accuracy, considers a prediction correct only when both outputs are correct. Our zero-shot experiments reveal a clear gap between the two objectives: GPT-5.6 achieves 91.53\% Answer Accuracy but only 74.58\% View Accuracy, resulting in 71.19\% Joint Accuracy. This suggests that improving grounded multi-view reasoning requires explicit mechanisms for verifying visual evidence rather than relying solely on stronger answer generation.

To address this issue, we propose \textbf{CoVeR-VQA}, a collaborative multi-stage verification and correction framework for grounded multi-view VQA. CoVeR-VQA starts from a zero-shot joint answer--view prediction produced by GPT-5.6 and subsequently applies three complementary verification stages for view auditing, joint correction, and cross-split group-level reasoning. Gemini-3.6-Flash first performs a dedicated audit of the predicted supporting view. Claude-Opus-5 then conducts confidence-aware joint verification of both the answer and the view. In the final stage, semantically filtered question groups across shared multi-view scenes are considered collectively to provide group-level prior knowledge for further verification and correction. When validation instances are available in the same group, their annotated answers and supporting views are additionally exploited as verification priors. This staged design targets different sources of error while reducing unnecessary changes to already reliable predictions. The resulting pipeline improves Joint Accuracy from 71.19\% to 84.75\%, with Answer Accuracy and View Accuracy reaching 94.92\% and 86.44\%, respectively. The final submitted run achieves 88.14\% Joint Accuracy on the official evaluator.

In summary, our contributions are:
\begin{itemize}
    \item We propose \textbf{CoVeR-VQA}, a training-free multi-stage framework that jointly improves answer prediction and view-level visual grounding in multi-view driving VQA.
    \item We introduce complementary verification and correction mechanisms that target view-selection errors, coupled answer--view errors, and prediction errors that can be resolved using semantically filtered group-level priors derived from related questions and, when available, validation instances within the same scene.
    \item We analyze the contribution of individual stages and show that explicit evidence verification substantially improves joint grounded reasoning over strong zero-shot MLLM baselines.
\end{itemize}

\section{Task and Data}

\subsection{Task Definition}

GoldenViewVQA~\citep{wang2026doesanswercomefrom} evaluates multi-view visual question answering with explicit evidence-source identification. For each instance, the model receives six synchronized camera observations from the NuScenes autonomous-driving dataset~\citep{caesar2020nuscenes}, corresponding to the front, front-left, front-right, back, back-left, and back-right views. Each instance additionally contains a natural-language question and four candidate answers. The system is required to jointly predict the answer and the camera view that provides the supporting visual evidence.

Formally, an input instance is represented as
\begin{equation}
x_i = \left(\mathcal{I}_i, q_i, \mathcal{A}_i\right),
\end{equation}
where $\mathcal{I}_i=\{I_i^1,\ldots,I_i^6\}$ denotes the six synchronized camera images, $q_i$ denotes the question, and $\mathcal{A}_i=\{A_i,B_i,C_i,D_i\}$ denotes the four answer candidates. Given the input, the system predicts
\begin{equation}
(\hat{v}_i,\hat{a}_i) = f(\mathcal{I}_i,q_i,\mathcal{A}_i),
\end{equation}
where $\hat{v}_i$ is the predicted supporting view and $\hat{a}_i$ is the predicted answer. The supporting-view prediction is selected from the six camera views or \texttt{NONE\_OF\_THE\_ABOVE}, while the answer prediction satisfies $\hat{a}_i\in\{A,B,C,D\}$.

The benchmark covers three categories of driving-scene reasoning: \emph{causality}, \emph{counterfactual reasoning}, and \emph{intent prediction}. Causality questions require identifying the visual factors that explain a driving decision or maneuver. Counterfactual questions ask how the situation would change under a hypothetical condition, while intent-prediction questions require inferring the likely future behavior of surrounding road users. Solving these questions often requires localized evidence involving traffic signals, pedestrians, road geometry, parked or moving vehicles, and other scene-specific constraints.

The primary evaluation metric is \emph{Joint Accuracy}, which considers an instance correct only when both the answer and the supporting view are predicted correctly:
\begin{equation}
    \mathrm{JointAcc}=\frac{1}{N}\sum_{i=1}^{N}\mathbf{1}\left[\hat{v}_i=v_i\land\hat{a}_i=a_i\right].
\end{equation}
In addition to Joint Accuracy, the official evaluator reports Answer Accuracy, View Accuracy, and View Macro Accuracy. These complementary metrics allow us to distinguish answer-generation errors from evidence-localization errors, which is particularly important because a model may produce the correct answer while grounding it in an incorrect camera view.

\subsection{Dataset}

\begin{figure*}[t]
    \centering
    \includegraphics[width=0.95\textwidth]{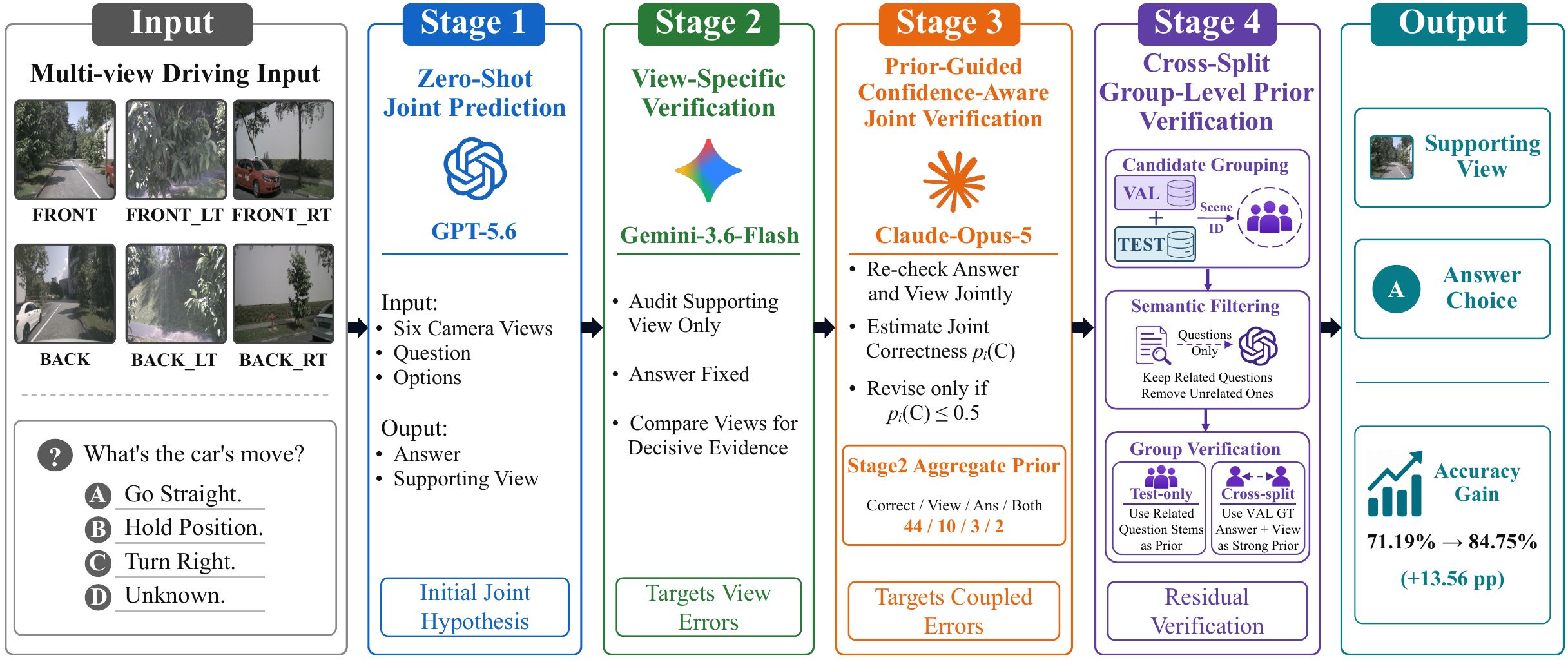}
    \caption{Overview of the proposed CoVeR-VQA framework and its four-stage prediction and verification pipeline.}
    \vspace{-4mm}
    \label{fig:system}
\end{figure*}

GoldenViewVQA is constructed from multi-view driving scenes in the NuScenes dataset~\citep{caesar2020nuscenes}. Each example contains six synchronized camera images together with a multiple-choice question, four candidate answers, and a supporting-view annotation. The shared-task release contains 114 instances in total, including 55 labeled validation examples and 59 unlabeled test examples, corresponding to 330 and 354 camera images, respectively. In the validation split, causality, counterfactual reasoning, and intent prediction account for 49.09\%, 27.27\%, and 23.64\% of the instances, respectively; the corresponding proportions in the test split are 35.59\%, 30.51\%, and 33.90\%. Across the complete shared-task release, causality accounts for 48 instances (42.11\%), while counterfactual reasoning and intent prediction each contain 33 instances (28.95\%).

For validation examples, both the correct answer and the supporting camera view are provided. The official test set follows the same input format, but its answer and supporting-view annotations are withheld; system predictions are evaluated through the GroundLM online evaluator. We use the public validation set to understand the task structure, examine common grounding ambiguities, and design the prompting and verification strategies used in CoVeR-VQA. Although the validation and test splits are evaluated separately, some instances correspond to the same underlying multi-view driving scenes. This cross-split scene relationship enables our Stage~4 verification strategy.

\section{CoVeR-VQA System}
\label{sec:cover}

As illustrated in Figure~\ref{fig:system}, CoVeR-VQA adopts a progressive multi-model verification framework for grounded multi-view visual question answering. Starting from a zero-shot joint answer--view prediction produced by GPT-5.6~\citep{openai2026gpt56}, the system sequentially applies view-specific verification with Gemini-3.6-Flash~\citep{google2026gemini36flash}, confidence-aware joint verification with Claude-Opus-5~\citep{anthropic2026opus5}, and cross-split group-level verification with GPT-5.6. Each stage receives the prediction from the preceding stage and targets a distinct source of error, including supporting-view mislocalization, coupled answer--view errors, and residual errors that can be identified through semantically related question groups and validation-derived prior knowledge.

\subsection{Stage 1: Zero-Shot Joint Prediction}

The first stage uses GPT-5.6 to obtain an initial joint prediction. For each instance, all six synchronized camera views are presented together with their camera identifiers, the question, and four candidate answers. The model is instructed to jointly determine the most appropriate supporting view and the corresponding answer choice.

We retain all six views during this stage rather than selecting a camera in advance. This allows the model to directly compare evidence across different viewing directions and reduces the risk of discarding useful information through premature view selection.

The resulting prediction is represented as
\begin{equation}
    P_i^{(1)} =\left(\hat{v}_i^{(1)}, \hat{a}_i^{(1)}\right),
\end{equation}
where $\hat{v}_i^{(1)}$ and $\hat{a}_i^{(1)}$ denote the predicted supporting view and answer, respectively.

Although this zero-shot setting already provides strong answer prediction, our experiments show that errors in supporting-view identification remain considerably more frequent. This discrepancy motivates the explicit view verification introduced in the next stage.

\subsection{Stage 2: View-Specific Verification}

The second stage focuses exclusively on the supporting-view prediction. Gemini-3.6-Flash receives the original six camera views, the question and answer candidates, as well as the answer--view pair produced by Stage~1.

Rather than solving the complete task again, the reviewer is instructed to determine whether the current supporting view actually contains the visual evidence required to justify the predicted answer. It compares the candidate views and re-evaluates relevant objects, spatial relations, and scene constraints.

Importantly, the answer produced in Stage~1 is kept unchanged during this stage. Only the supporting-view prediction is allowed to be corrected:
\begin{equation}
    P_i^{(2)}=
    \left(
        \hat{v}_i^{(2)},
        \hat{a}_i^{(1)}
    \right).
\end{equation}
This design is motivated by the observation that a model may already produce the correct answer while grounding it in an incorrect camera view. Restricting Stage~2 to view verification therefore reduces the risk of replacing an already-correct answer while correcting the grounding decision.

\subsection{Stage 3: Prior-Guided Confidence-Aware Joint Verification}

Although Stage~2 improves supporting-view localization, it cannot correct cases in which the answer itself is incorrect or both components of the prediction are unreliable. We therefore introduce a Claude-based joint verification stage that simultaneously re-examines the answer and supporting view.

After Stage~2, the official evaluation metrics provide additional information about the aggregate error structure of the current predictions. Specifically, among the 59 test instances, 44 predictions are jointly correct, 10 contain only a view error, 3 contain only an answer error, and 2 contain errors in both components. We treat this aggregate distribution as \emph{Stage-2-derived prior knowledge}. Importantly, this prior describes only the error distribution of the entire test set and does not reveal which individual instances belong to each category.

For each instance, Claude receives the six synchronized camera views, the question and answer candidates, the current Stage~2 prediction
$P_i^{(2)}$, and the Stage-2-derived prior.

Claude independently audits the current prediction and estimates its likelihood under four possible states:
\begin{equation}
    \mathcal{Z}=\{\mathrm{C},\mathrm{V},\mathrm{A},\mathrm{B}\},
\end{equation}
where $\mathrm{C}$ indicates that both the view and answer are correct, $\mathrm{V}$ represents a view-only error, $\mathrm{A}$ represents an answer-only error, and $\mathrm{B}$ indicates that both predictions are incorrect. The reviewer also proposes corresponding correction candidates when it identifies potential errors.

The aggregate prior is used only to provide the reviewer with information about the overall reliability and error composition of the Stage~2 predictions. We do not enforce the 44/10/3/2 distribution as a hard assignment over individual instances. Instead, the final correction decision is determined by Claude's instance-level confidence.

Let $p_i(\mathrm{C})$ denote Claude's estimated probability that the current answer--view pair is jointly correct. To avoid unnecessarily altering reliable predictions~\citep{hu2026glance,liu2026interference}, we apply a conservative confidence threshold and revise only instances for which
\begin{equation}
    p_i(\mathrm{C}) \leq 0.5.
\end{equation}
The Stage~3 prediction is therefore defined as
\begin{equation}
    P_i^{(3)}=
    \begin{cases}
        \tilde{P}_i^{(3)},
        & p_i(\mathrm{C}) \leq 0.5, \\[2pt]
        P_i^{(2)},
        & p_i(\mathrm{C}) > 0.5,
    \end{cases}
\end{equation}
where $\tilde{P}_i^{(3)}$ denotes the correction proposed by Claude after jointly reconsidering the supporting view and answer.

This design combines collection-level prior knowledge with instance-level multimodal verification. The prior informs Claude that most Stage~2 predictions are already reliable and provides information about the relative prevalence of different error types, while the confidence threshold prevents this aggregate information from forcing a predetermined number of modifications. As a result, Stage~3 selectively targets predictions for which the reviewer finds stronger evidence against the current answer--view pair~\citep{liu2026chartlens,yu2026parsefixer}.

\subsection{Stage 4: Cross-Split Group-Level Prior-Guided Verification}

The previous stages process each question largely independently, although multiple instances may correspond to the same multi-view scene and involve related entities or spatial relations. Stage~4 therefore introduces group-level verification by exploiting semantically related questions across both the validation and test splits.

We first merge the validation and test instances and construct candidate scene groups by exploiting the shared base \texttt{sfall\_XXXX} identifier together with synchronized camera observations. Since questions from the same scene may still refer to different objects or require different evidence, we further apply GPT-5.6 as a text-only semantic filter. The model receives only the question stems and retains questions in the same group when they either concern the same core entity and local spatial relation, or exhibit a sufficiently strong semantic, causal, or behavioral connection within the scene. Semantically unrelated questions are separated, and singleton groups are discarded. The resulting group set is
\begin{equation} \mathcal{G}=\{G_1,G_2,\ldots,G_M\}, \qquad |G_m|\geq 2. \end{equation}

For each retained group, GPT-5.6 is provided with the shared six-view scene, the questions, answer options, available annotations of group members, and the current predictions from Stage~3. It then re-evaluates the answer and supporting view of each test instance. We denote the textual group prior by $\Pi_{\mathrm{text}}(G_m)$ and the validation-derived prior by $\Pi_{\mathrm{val}}(G_m)$. For test-only groups, $\Pi_{\mathrm{val}}(G_m)=\emptyset$, whereas for cross-split groups it contains the ground-truth answers and supporting views of the related validation instances. The complete group-level prior is therefore
\begin{equation}
\Pi(G_m)=\Pi_{\mathrm{text}}(G_m)\cup\Pi_{\mathrm{val}}(G_m).
\end{equation}

Validation annotations provide a particularly strong reference when the validation and test questions concern the same core entity or local spatial configuration. In such cases, GPT-5.6 is encouraged to maintain supporting-view consistency with the validation instances. This consistency is not enforced as a hard constraint, however, since different questions from the same scene may require different decisive evidence. A different view may therefore be selected when supported by clear question-specific visual evidence.

For a test instance $i$ belonging to group $G_m$, Stage~4 jointly conditions on the original input $x_i$, the previous prediction $P_i^{(3)}$, and the corresponding group-level prior $\Pi(G_m)$:
\begin{equation} P_i^{(4)}=f_{\mathrm{group}}\left(x_i, P_i^{(3)},\Pi(G_m)\right). \end{equation}

By combining semantic group filtering with validation-derived cross-split priors, Stage~4 uses related questions as targeted prior knowledge while avoiding indiscriminate information transfer between questions that merely share the same scene. This design is particularly useful for correcting residual answer--view inconsistencies that are difficult to resolve from individual questions alone.

\section{Experimental Results}

\subsection{Experimental Details}

Our experiments involve four multimodal large language models: Qwen3-VL-8B~\citep{bai2025qwen3vl}, GPT-5.6~\citep{openai2026gpt56}, Gemini-3.6-Flash~\citep{google2026gemini36flash}, and Claude-Opus-5~\citep{anthropic2026opus5}. Qwen3-VL-8B is used as an open-model zero-shot baseline, while GPT-5.6, Gemini-3.6-Flash, and Claude-Opus-5 are used at different stages of the CoVeR-VQA pipeline. We do not perform task-specific fine-tuning or use additional labeled training data; all improvements are obtained through inference-time prediction, verification, and correction. Detailed prompts are provided in the accompanying code submission.

Stage~3 uses only aggregate evaluator statistics from the complete Stage~2 submission as a collection-level prior and applies a fixed confidence-based verification procedure to all instances. Stage~4 further uses semantically filtered relationships between validation and test instances. Validation annotations are only introduced within cross-split groups identified by shared scenes and semantic relevance. The resulting four-stage CoVeR-VQA pipeline achieves 84.75\% Joint Accuracy on the official GoldenViewVQA evaluator. For the final submission, we additionally examined two instances that Claude identified as potentially erroneous but that did not satisfy the correction threshold. Based on subsequent evaluator feedback, we applied two instance-level post-hoc corrections, increasing Joint Accuracy to 88.14\%. Since these corrections do not constitute a generalizable inference component, we report the final submission separately from the four-stage pipeline. The two corrected instances are revisited in Section~\ref{sec:error-analysis} as representative pre-correction failure cases.

\subsection{Overall Performance}

Table~\ref{tab:official-results} summarizes the performance of the evaluated systems on the GoldenViewVQA test set. The organizer baseline achieves only 20.34\% Joint Accuracy despite a relatively high View Accuracy of 77.97\%, indicating that accurate view prediction alone is insufficient without reliable answer reasoning. The zero-shot Qwen3-VL-8B baseline substantially improves Joint Accuracy to 66.10\%, while GPT-5.6 further raises it to 71.19\%, together with 91.53\% Answer Accuracy and 74.58\% View Accuracy.

The complete four-stage CoVeR-VQA pipeline reaches 84.75\% Joint Accuracy, improving the GPT-5.6 zero-shot baseline by 13.56 percentage points. Answer Accuracy increases from 91.53\% to 94.92\%, while View Accuracy improves from 74.58\% to 86.44\%. These results show that the proposed verification stages improve not only answer correctness but, more importantly, the consistency between the predicted answer and its supporting visual evidence.

\begin{table}[t]
    \centering
    \small
    \setlength{\tabcolsep}{3.2pt}
    \begin{tabular}{lcccc}
        \toprule
        \textbf{Method} & \textbf{Joint} & \textbf{Ans.} & \textbf{View} & \textbf{V-Macro} \\
        \midrule
        
        Organizer Baseline
        & 20.34 & 23.73 & 77.97 & 16.67 \\
        
        Qwen3-VL-8B Zero Shot
        & 66.10 & 88.14 & 71.19 & 49.88 \\
        
        \rowcolor{gray!10}
        \stage{1} GPT-5.6 Zero Shot
        & 71.19 & 91.53 & 74.58 & 65.82 \\
        
        + GPT-5.6 View Review
        & 72.88 & 91.53 & 77.97 & 64.13 \\
        
        \rowcolor{gray!10}
        \stage{2} Gemini View Review
        & 74.58 & 91.53 & 79.66 & 54.11 \\
        
        \rowcolor{gray!10}
        \stage{3} Claude Joint Review
        & 79.66 & 93.22 & 83.05 & 54.83 \\
        
        \rowcolor{gray!10}
        \stage{4} Cross-Split Group Verify
        & 84.75 & 94.92 & 86.44 & 65.94 \\
        
        \midrule
        
        \textbf{CoVeR-VQA Pipeline}
        & \textbf{84.75}
        & \textbf{94.92}
        & \textbf{86.44}
        & \textbf{65.94} \\
        
        \textbf{Final Submission}
        & \textbf{88.14}
        & \textbf{96.61}
        & \textbf{89.83}
        & \textbf{74.64} \\
        
        \bottomrule
    \end{tabular}
    \caption{Results on the GoldenViewVQA test set (\%). Shaded rows marked with \stage{1}--\stage{4} denote the four stages of the CoVeR-VQA pipeline, while ``+'' denotes an additional ablation setting.}
    
    \vspace{-4mm}
    \label{tab:official-results}
\end{table}
\begin{figure}[t]
    \centering
    \includegraphics[width=0.9\columnwidth]{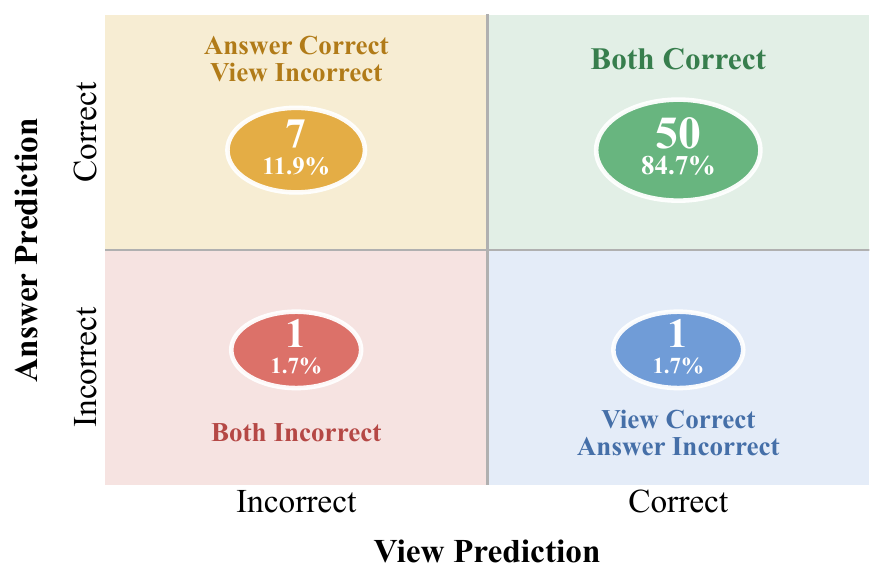}
    \vspace{-2mm}
    \caption{Distribution of answer and supporting-view correctness for the final submission on the GoldenViewVQA test set.}
    \vspace{-4mm}
    \label{fig:view-answer-quadrant}
\end{figure}

The final submitted system achieves 88.14\% Joint Accuracy, 96.61\% Answer Accuracy, and 89.83\% View Accuracy. As illustrated in Figure~\ref{fig:view-answer-quadrant}, 52 of the 59 test instances are correct in both answer and view prediction. Among the remaining seven instances, five have the correct answer but an incorrect supporting view, while only one instance contains an answer-only error and one contains errors in both components. This error distribution further suggests that view localization remains the dominant source of residual error after answer prediction has largely been resolved.

\begin{figure}[t]
    \centering
    \includegraphics[width=0.95\columnwidth]{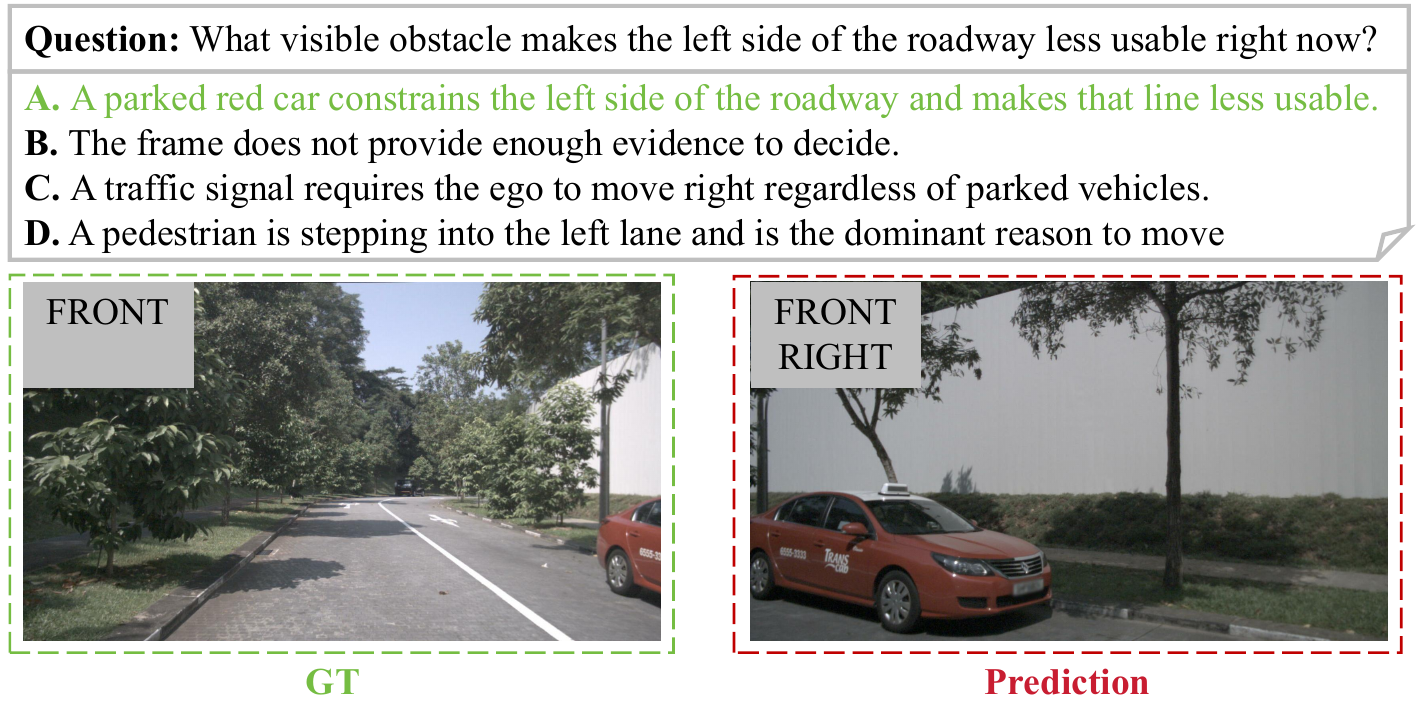}
    \vspace{-2mm}
    \caption{Example of a view-selection error caused by confusing object saliency with decisive spatial evidence.}
    \vspace{-4mm}
    \label{fig:error1}
\end{figure}

\subsection{Stage-wise Analysis and Ablations}

The stage-wise results in Table~\ref{tab:official-results} reveal complementary effects across the CoVeR-VQA pipeline. Compared with the GPT-5.6 self-review ablation, Gemini-3.6-Flash increases View Accuracy from 77.97\% to 79.66\% and Joint Accuracy from 72.88\% to 74.58\%, while Answer Accuracy remains unchanged at 91.53\%. This comparison confirms the effectiveness of the Stage~2 review mechanism and further indicates that cross-model verification is more effective than self-review with the same model for supporting-view correction.

The largest improvement is obtained at Stage~3. Claude-based joint verification raises Joint Accuracy from 74.58\% to 79.66\%, with simultaneous gains in Answer Accuracy and View Accuracy from 91.53\% to 93.22\% and from 79.66\% to 83.05\%, respectively. This result supports the motivation for jointly reconsidering answer and view predictions after view-specific auditing, as some remaining errors cannot be resolved by modifying only the supporting view.

Stage~4 further improves Joint Accuracy to 84.75\% and Answer Accuracy to 94.92\%, while View Accuracy further increases to 86.44\%. This suggests that group-level verification provides complementary information for correcting residual prediction errors, where related question contexts offer semantic priors and validation annotations provide stronger evidence-aware guidance when available. Notably, View Macro Accuracy does not increase monotonically across stages, indicating that improvements in overall view correctness are not uniformly distributed across all view categories.

Overall, the ablation results show that the gains of CoVeR-VQA arise from complementary verification mechanisms rather than repeated applications of the same reviewer. View-specific auditing primarily improves grounding, joint verification corrects coupled answer--view errors, and group-level prior knowledge provides additional gains for difficult residual cases.

\subsection{Error Analysis}
\label{sec:error-analysis}

A major source of error is the confusion between \emph{object visibility} and \emph{decisive spatial evidence}. As illustrated in Figure~\ref{fig:error1}, the red parked vehicle is more visually prominent in the front-right camera, which can easily attract the model toward that view. However, the task does not ask which camera provides the clearest or largest appearance of the target object; it asks which view contains the evidence needed to establish the relevant spatial relation. In this example, the front view better reveals the position of the parked vehicle with respect to the ego lane and the usable roadway, thereby supporting the conclusion that the left side is constrained. This failure suggests that supporting-view selection should be based on the complete action-relevant scene configuration---including object position, lane geometry, ego orientation, and their spatial relations---rather than object saliency alone~\citep{li2025dcount,wang2026visage}. In other words, the golden view is the view that provides the decisive evidence for answering the question, not necessarily the view in which the key object appears most prominently.

\begin{figure}[t]
    \centering
    \includegraphics[width=0.9\columnwidth]{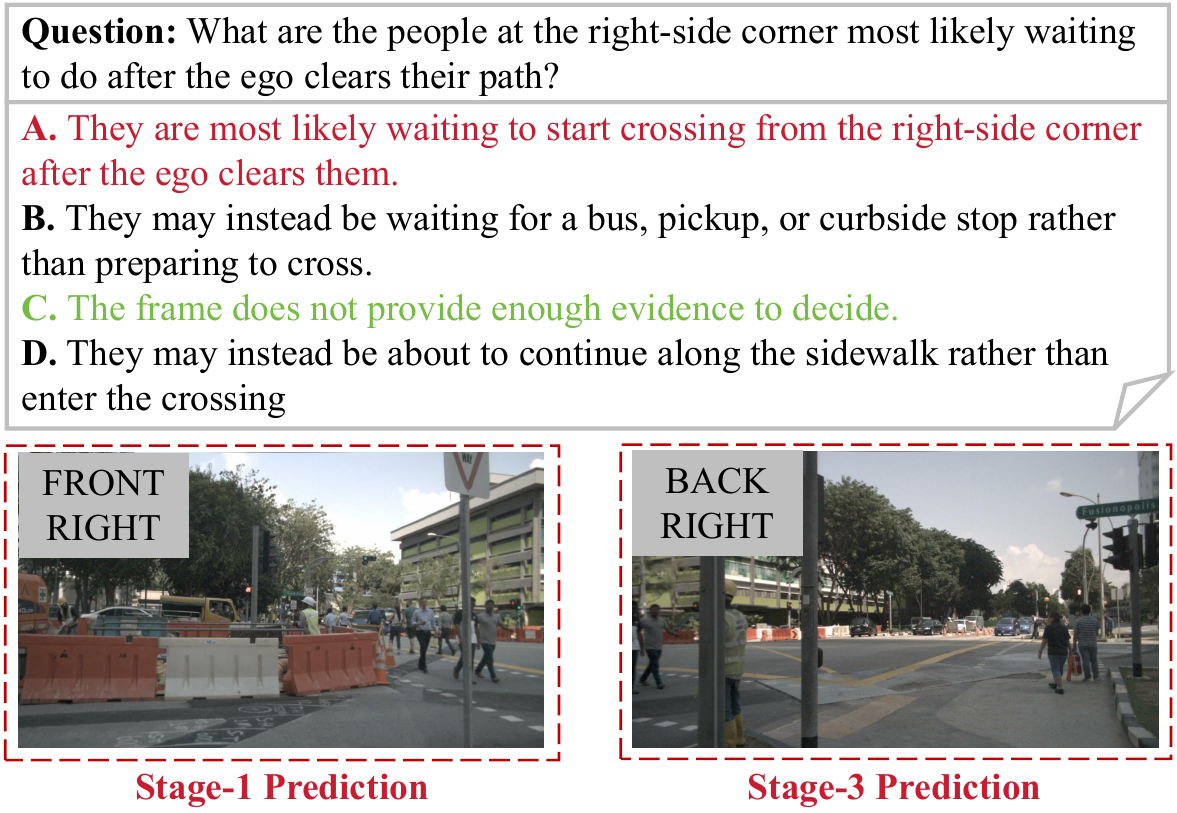}
    \vspace{-2mm}
    \caption{Example of unsupported semantic inference under ambiguous visual evidence.}
    \vspace{-4mm}
   \label{fig:error2}
\end{figure}

A second recurring failure mode is \emph{unsupported semantic inference}. Figure~\ref{fig:error2} shows an intent-prediction example in which pedestrians are visible near the right-side corner, but the available views do not provide sufficient evidence to determine whether they intend to cross, wait for transportation, or continue along the sidewalk. Nevertheless, the model tends to select a visually plausible view and infer a likely behavior instead of recognizing the absence of decisive evidence. Such behavior reflects a form of multimodal hallucination, a known failure mode of large vision-language models \citep{li2023pope, guan2024hallusionbench}: when visual evidence is ambiguous or incomplete, the MLLM may fill the missing information with a semantically plausible interpretation derived from common driving scenarios. This is particularly problematic for GoldenViewVQA, where \texttt{NONE\_OF\_THE\_ABOVE} is a valid supporting-view label and some answer choices may explicitly indicate that the available evidence is insufficient. Effective grounded reasoning therefore requires not only identifying supporting evidence when it exists, but also calibrating uncertainty and refraining from committing to an answer--view pair when the observations do not provide sufficient support.

\section{Conclusion}
We presented \textbf{CoVeR-VQA}, a training-free multi-stage framework for joint answer prediction and view-level evidence grounding in GoldenViewVQA. By progressively combining zero-shot prediction, view-specific auditing, prior-guided joint verification, and cross-split group-level verification with semantic filtering and validation-derived priors, CoVeR-VQA improves Joint Accuracy from 71.19\% to 84.75\%, with the final submitted run reaching 88.14\%. Stage-wise analysis demonstrates that the verification components provide complementary gains in answer correctness and supporting-view localization. Error analysis further reveals that the remaining difficulty lies primarily in identifying decisive spatial evidence rather than salient objects, as well as in avoiding unsupported predictions when visual evidence is insufficient. These findings suggest that reliable grounded multimodal reasoning requires not only stronger semantic reasoning, but also explicit evidence verification and better uncertainty calibration.

\section*{Limitations}

Our approach has several limitations. First, the public validation set contains only 55 labeled instances, which limits systematic prompt optimization and error analysis and may make observations derived from the validation data sensitive to individual examples.

Second, CoVeR-VQA remains strongly dependent on the underlying reasoning, calibration, and behavioral reliability of MLLMs~\citep{li2026mist}. As discussed in our error analysis, current MLLMs may produce plausible interpretations even when the available visual evidence is insufficient, rather than explicitly selecting \texttt{NONE\_OF\_THE\_ABOVE}. Such hallucination behavior remains a major challenge for reliable evidence-grounded reasoning, while residual vulnerabilities under model-level interventions represent a broader reliability concern~\citep{xiang2026tina,xiang2025dkdm}.

Finally, Stage~4 relies on semantically related question groups within shared multi-view scenes and, for cross-split groups, on annotations from related validation instances. This strategy exploits structural relationships available in the released benchmark and is therefore specific to settings where such cross-instance relationships are accessible. These conditions are generally unavailable in standard independent-test or real-time autonomous-driving scenarios, limiting the general applicability of Stage~4 beyond the shared-task setting.

\bibliography{custom}

\end{document}